# Taylor-based Optimized Recursive Extended Exponential Smoothed Neural Networks Forecasting Method

Emna Krichene[a], Wael Ouarda[a], Habib Chabchoub[b], Adel M. Alimi[a]

[a]REGIM-Lab:REsearch Groups in Intelligent Machines, University of Sfax, National Engineering School of Sfax (ENIS),BP 1173, Sfax 3038, Tunisia [b]GIAD Unit research, FSEGS, University of Sfax, Tunisia emna.krichene.tn@ieee.org, wael.ouarda@ieee.org, habib.chabchoub@fsegs.rnu.tn, adel.alimi@ieee.org

---

Abstract

A newly introduced method called Taylor-based Optimized Recursive Extended Exponential Smoothed Neural Networks Forecasting method is applied and ex-tended in this study to forecast numerical values. Unlike traditional forecasting techniques which forecast only future values, our proposed method provides a new extension to correct the predicted values which is done by forecasting the estimated error. Experimental results demonstrated that the proposed method has a high accuracy both in training and testing data and outperform the state-of-the-art RNN models on Mackey-Glass, NARMA, Lorenz and Henon map datasets.

---

## 1. Introduction

Nowadays, data are increasingly growing. Hence, data computation becomes a major challenge. Companies dealing with Big Data need sophisticated decision assistance in order to optimize the planning of these tasks and tools which have got several problems to achieve their goals. First, the heterogeneity of the data

---

Corresponding author

Email address: emna.krichene.tn@ieee.org (Emna Krichene)

induces the complexity of forecasting future values. Second, the diversity of the methods involved in the computation (i.e., in addition to the several inherent constraints) increases this complexity.

Currently, an important aspect related to forecasting represents the subject of much researches. Indeed, it is possible to observe accurately the current status of dataset as well as their historic and subsequently to evaluate various parameters. However, it is possible to go further than the simple information extraction, and to move towards real predictive analytics information.

Forecasting is an important data analysis eld that aims to examine historical data in order to extend and predict its future values. Thus, forecasting activities play an important role in our daily life and involve various elds so that much research aim to develop tools for forecasting and decision making. We often forecast weather [1][2][3][4], wind speed [5] [6], word speech recognition [7], image recognition [8][9][10][11][12], face recognition[13], writer identi cation [14], stock market [15], electricity [16][17][18][19][20], etc.

Meanwhile, forecasting literature contains a wide diversity of techniques that can be organized into two main families [20] [21]:

(i) Qualitative methods [22][23][24] where judgmental forecasts are princi-pally based on experts' opinions. These methods provide in forecasting any information that cannot be explained in the past. Furthermore, these techniques are, especially, used when su cient information and data are not available; so that quantitative methods cannot be used i.e., when case study is vague and only few data exist. For example, bearing on appari-tion of a new product or a new event, this type of forecasting techniques seems very adequate. The most known qualitative methods are: Con-sumer Survey [25], Consumer Survey-Sample Survey Method [25], Delphi Method [26][27] and Past Analogies [28]. However, in spite of their ro-bustness, these methods present some disadvantages. As they are totally based on the intuitions of experts, these techniques are not only expensive but also not adequate for most cases especially in big data case; i.e., hu-man capability is not able to predict future values from a wide range and observations. For these reasons and others, another type of forecasting techniques appears which is totally based on mathematical fundamentals later called as quantitative methods.

(ii) Quantitative methods (also called statistical methods) [29][30][31][32] do not relate to experts' intuitions but they mostly rely on quantitative his-torical data that can be extrapolated to make our forecasts. This type of techniques is used when the case study is stable and historical data exist. The most known statistical techniques are: Naive method [33], Moving Average (MA) [34], Weighted Moving Average (WMA) [34], Exponential Smoothing [35] and linear regression method [36]. These techniques are commonly used when they handle products or phenomena that already exist and when the historical data are of course available in order to study it. Comparing to qualitative methods, these techniques are less expensive and faster but their e ectiveness is limited to some applications as they still be not appropriate when observations are very di erent and chaotic. This point motivated researchers to go even further and to move toward arti cial intelligence as prediction tools.

Recently, di erent methods have been presented to the task of forecasting as intelligent techniques. Neural Networks (NN) are the most-used techniques as they are nonparametric and nonlinear methods. NNs are able to model and to map complex problems by training them in order to determine parameters and explore relationships between data [37][38][39]. Accordingly, NNs have been considered as a promising and useful method applied to a variety of forecasting problems.

From an architectural point of view, there are various types of neural net-works that share the same basic concept. Each problem needs to be molded by admitting its own neural network architecture and network con guration since there is no single learning model that can be suitably used for all elds. However, in NN, we distinguish two main categories: (i) Multi-Layer percep-tron (MLP) which is the most popular method applied in time series forecasting task and (ii) Recurrent NN (RNN) [37][40][41] where its context-layer serves to store previous states within a dynamic memory. Authors in [39] believe that the interest of this supplementary layer is its ability to better underline the relationships between past observations and future values and so that a better generalization is enhanced.

A brief comparison of these techniques is summarized and presented in table 1.

Table 1: FORECASTING TECHNIQUES COMPARISON

| Family of the technique | Description (main concept and advantages) |
|---|---|
| Qualitative methods [22][23][24][25][26][27][28] | - Depend upon experience of individual experts.<br>- Suitable for new phenomena, product, event and when few data exists.<br>- Not able to model complex data especially in big data case. |
| Quantitative methods [29][30][31][32][33][34][35][36] | - Rely on mathematical formulas<br>- Suitable when historical data exist.<br>- Not able to model complex data. |
| Intelligent methods [40][41][42][43] | - Rely on intelligent theories.<br>- Suitable for complex and chaotic data.<br>- Capability to map the relationship between data. |

This paper focuses on intelligent methods and examines the potential of a family of arti cial neural networks called Recurrent Neural networks (RNN). Three interesting goals motivated us to propose this new architecture: First, unlike feed-forward neural networks, RNN have rather received less attention in the forecasting literature. Second, theoretically, most of forecasting methods assume that a fundamental relationship exists between the historical observations of a time series and its future values. Due to the complexity of the task, traditional forecasting techniques seem not able to estimate correctly this underlying relationship. Thanks to their distinct structure, RNN can be a good model to satisfy this assumption as it is well suited to retain past observations in internal memory (context units) to use them during the training phase. Finally,

RNNs are nonlinear; indeed, real world systems are repeatedly nonlinear [39], among others the Mackey-Glass time series, Lorenz benchmark and Henon map already experimented in this study.

In this context, we draw inspiration from the Exponential Smoothed method (ES) [35]. Thereby, Recursive Extended Exponential Smoothed Neural Networks (REESNN) represents the rst part conception of our proposed method. The primary objective of this work is an assessment of this method as a tool of forecasting. Furthermore, a secondary part of this paper consists on optimizing the results already obtained in the rst step. The task of optimization is per-formed by predicting the estimated error to be then tailored with forecasting results by applying Taylor expansion principle.

This paper is organized as follows: In Section 2, we start by a review in which we will discuss recent researches using Recurrent Neural network architectures. Then, Section 3 gives an extensive description of the propounded approach based on Taylor theorem to optimize forecasting future values. Then, we will illustrate the robustness of our approach by giving an exhaustive experimental result on four di erent datasets: Mackey-Glass times series, NARMA, Lorenz and Henon-Attractor in Section 4. Finally, Section 5 draws the conclusion of the paper.

## 2. Related works

Nowadays, a signi cant aspect related to recurrent neural network (RNN) performs the matter of much forecasting research. The literature is vast and growing and it is di cult for a researcher to sweep all works contributing to forecasting literature done until now. In this section, we will present some works done over the last ten years that apply RNN as forecasting techniques. Hence, di erent architectures of RNN have been studied. However it seems that the most in uential models are the Elman RNN (ERNN) [40] and Jordan RNN (JRNN) [43] as they are the most used models.

This section is divided into two parts: In the rst one, we explain the funda-mental principle of ERNN and we present some works contributing with ERNN architecture. In the second part, we brie y describe the architecture of Jor-dan RNN and we introduce a brief survey of research activities contributing to forecasting literature using Jordan RNN over the last 10 years.

### 2.1. Elman Recurrent Neural Networks

The simplest and the most known Recurrent Neural Network is ERNN architecture [40]. ERNN is designed as three layers expanded by a context-layer which receives inputs from the hidden units (see Figure 1). The role of this supplementary layer is to store previous states of the hidden neurons within a dynamic memory.

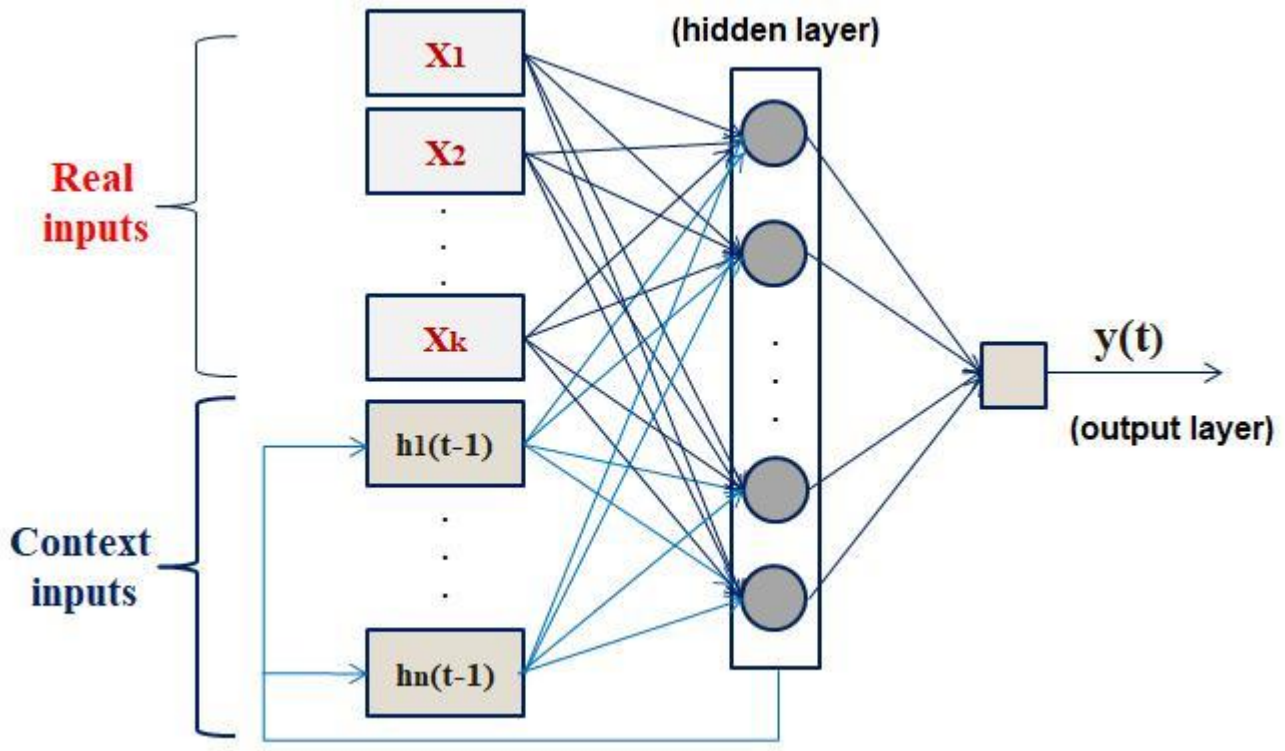

Figure 1: The architecture of Elman RNN model

ERNN is the most popular and widely-used technique on Arti cial Intelligence applications. Nevertheless, many researchers use ERNN as forecasting tool. Among them, authors in [15] applied four di erent types of RNN for sales forecasting (i.e., Elman RNN, Jordan RNN, Hierarchical Elman recurrent neural network (HERNN) and Multi-recurrent network). Then, they propose to analyze dissimilarities between the obtaining results and other quantitative forecasting methods (i.e., Exponential smoothing method [35] and SARIMA method [44]). Empirical results provide that HE outperforms other types of recurrent neural networks, whereas it is noticed that the Exponential method accurate better on longer horizon than other methods.

In the same context, researchers in [45] proposed to apply ERNN and Multi-Layer Perceptrons (MLP) as forecasting techniques to be later compared. Their study shows that both networks give good predictions but it can be seen that Elman recurrent neural network was more accurate than the multi-layer per-ceptron.

In [46], two di erent architectures of Elman RNN was applied as a predictor method to the consumer price index (CPI) of education, recreation, and sports data in Yogyakarta. Both architectures applied the sigmoid bipolar function within the hidden layer, whereas the transfer function within the output layer is the linear function. Every model was trained ve times using di erent number of hidden neurons (i.e., 1; 2; 3; 4 and 5 neurons), thus, their architectures were the same except that the number and type of input units was di erent. Hence, experiment results show that the design of the rst model accurate better than the second one not only during the training step but also during the testing one.

In the same context, authors in [47] applied the Elman recurrent neural network in order to classify recorded electroencephalogram (EEG). The speci city of this architecture is the use of the variant of hyperbolic tangent as an activation function within the hidden layer. We note that this transfer function was rst proposed by LeCun et al. in [48].

Another application of Elman recurrent neural network contributes to the literature of forecasting was applied by [2]. The authors employ di erent RNN architectures, di erent algorithms and various transfer functions in order to select the optimized model, i.e the model that gives accurate results.

In the same sens, authors in [3] applied also the Elman recurrent neural network architecture to predict the speci c humidity and compare the obtained results to those calculated with the multilayer perceptron network. Experiments prove once again that the recurrent topology outperforms the MLP. The authors have mentioned that the Elman RNN may be as accurate as MLP except in the case of large quantity of noise is restrained in the data.

Analogously, researchers in [16] design an Elman recurrent neural network to forecast electricity short term load. After trials, the architecture is designed as follows: four past inputs, ten units within the hidden layer and only one unit in the output layer. Then, the authors aim to add an exogenous factor so that the proposed latter architecture was changed to 8 10 1. The sigmoid function was implemented in both models with the hidden layer whereas the linear function was applied in the output layer.

Since Elman RNNs have a high performance, once again another work adopts its architecture to predict the electric load. Thereby, authors in [17] used two types of Elman RNN: the rst one with single hidden layer and the second with dual hidden layers. Experimental results show that the rst typology is more accurate as the average of the smallest error was given by it.

In [18], Elman RNN was also applied to forecast the electricity market price. The gradient descent back propagation algorithm was used during the learning phase in order to settle the parameters and building the optimized architecture. The hidden layer and output layer used the hyperbolic and linear function respectively to activate the neurons. A comparative study was then performed to di erent statistical models and shows that the recurrence accurate better in most cases.

However, based on the Jordan recurrent neural network architecture, many other papers contribute to the literature. In this section and the rest of the paper, we focus our interest on the Jordan RNN model and its variations.

## 2.2. Jordan Recurrent Neural Networks

The major di erence between Jordan RNNs architecture and those of other RNNs is that the context-layer is designed as a copy of the output-layer as shown in Figure 2.

The Jordan RNN [43] models are used in diverse prediction tasks especially in time series forecasting problems because of their inherent competence (effectiveness) to map input-output problems. In [49], authors build a variant architecture of Jordan RNN using exogenous input to forecast wave data; experimental results show that the proposed architecture presents more de nite results when the input data has been smoothed; applying the smoothing is two-fold: rst to eliminate the noise from the data, and second to give more importance to recent observations.

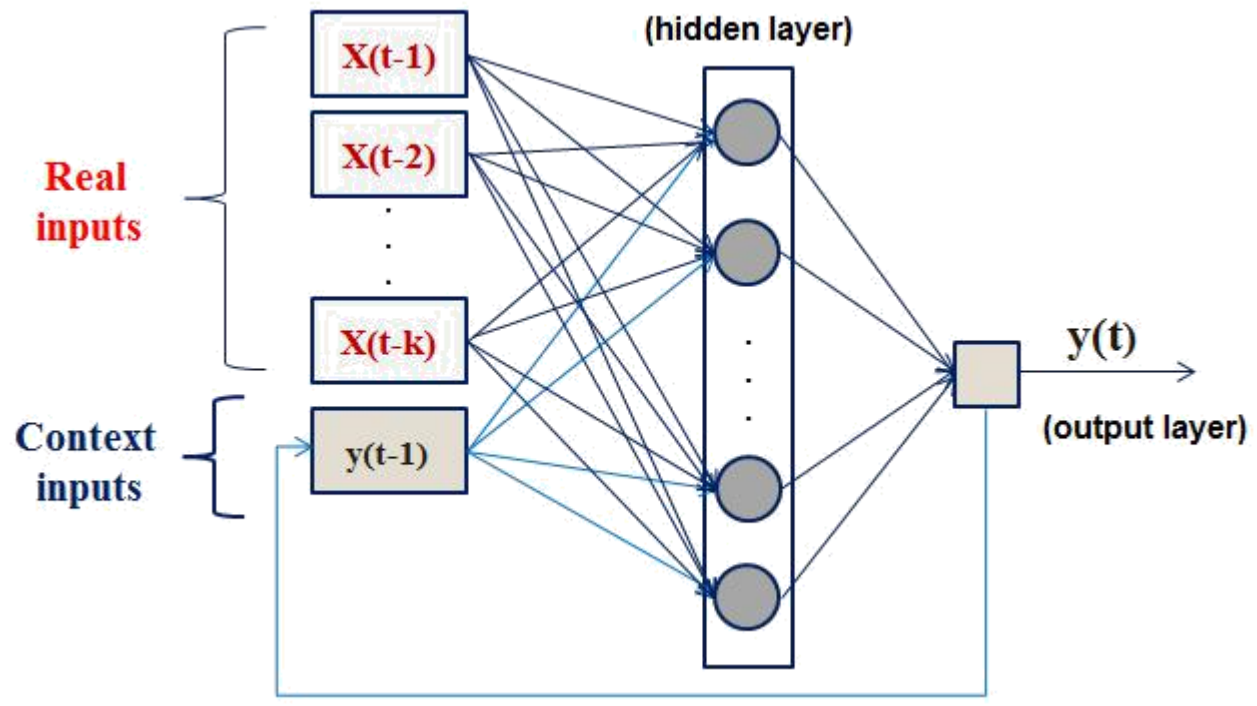

Figure 2: The architecture of Jordan RNN model

In the same sense, another forecasting problem has been solved by Jordan RNN. Authors in [5] applied two di erent architectures of Jordan recurrent neu-ral networks in order to analyze the wind speed prediction. Their study com-pares the proposed scheme with ARIMA model and the testing results con rms the assumption that recurrent neural network outperforms in most cases others statistical models. Another important application of recurrent neural network was proposed by [19], where the authors proposed a variation of Jordan recur-rent neural network called Recurrent Cartesian Genetic Programming evolved Arti cial Neural Network (RCGPANN). Comparing to other previous studies, the evaluation results show that the implemented method accurate better.

In [50], researchers present a new Jordan RNN model based on the Hy-brid Complex neural network; the proposed architecture is built as a recurrent multilayer perceptron as it has two hidden layers in addition to the recurrent connections. Results were then compared to those of feed-forward architecture and fully RNN model and it was noticed that the proposed scheme outperforms their results even in the worst case. Moreover, authors in [51] applied the Jor-dan RNN and have compared it to the MLP architecture. Thereby, although testing results were very similar, it can be noticed that the best forecasts were given by the recurrent architecture.

Likewise, Cappizi etal. [52] proposed a novel technique called wavelet RNN in which its principle is inspired from the Jordan RNN. Brie y describing, the proposed architecture is a multilayer recurrent network with output feedback connection to the rst hidden layer. A brief recapitulation of these contributing papers is summarized and presented in table 2.

Nevertheless, thanks to their speci c design, Jordan RNN is supposed to be a perfect alternative model to forecast time series even if their elements are hard to predict. Its structure is supposed to have a more powerful competence than other typical RNN architectures to train and predict the given data [46].

In this paper, to address the aforementioned problem, we propose a newly Recurrent NN architecture inspired from Taylor theory as a forecasting technique. Speci cally, we have aligned our contribution by focusing on two points: (1) Function approximation using the Recursive Extended Exponential Smoothed method and (2) error approximation in order to optimize the forecasted values using the Taylor expansion [53].

## 3. Proposed Approach

The proposed methodology is divided into three main steps given as follow:

Step 1: The proposed method Recursive Extended Exponential Smoothed Neural Networks (REESNN) inspired from the Jordan RNN & ES is applied in order to approximate the function and forecast values.

Step 2: Elman RNN Classi er is trained to approximate the forecasted error.

Step 3: The nal forecasting results are obtained by tailoring the two previous results according to the principle of Taylor expansion.

These aforementioned steps are more detailed in next sections. The fore-casting owchart of the proposed methodology is shown in Figure 3.

Table 2: Summary of di erent contributing papers over last Ten years

| | Reference+date | Data | Brief Description |
|---|---|---|---|
| ELMAN RNN | [2] (2010) | Sunspot numbers | Forecasting sunspot numbers using di erent architecture of ERNN. |
| | [16] (2010) | Electricity | Forecasting Electricity short term load using ERNN with exogenous factor. |
| | [47] (2011) | EEG | Classi cation of recorded electroencephalogram |
| | [17] (2012) | Electricity | Comparison between two architectures of ERNN: single hidden layer and dual hidden layers to forecast electric load. |
| | [18] (2012) | Electricity Market price | Comparison between ERNN and other intelligent techniques to forecast electricity market price. |
| | [45] (2012) | Mackey-Glass time series | Comparison between ERNN and MLP forecasting results. |
| | [3] (2014) | Speci c Humidity | Comparison between MLP and ERNN to forecast speci c humidity. |
| | [46] (2014) | Consumer price index of education | Comparison between two di erent architectures of ERNN. |
| | [15] (2015) | real-world sales data | Partial RNN was applied to prove the e ectiveness of PRNN in business decision support and sales forecasting. |
| JORDAN RNN | [49] (2010) | Wave data | A variant of Jordan RNN with exogenous data to forecast wave data. |
| | [50] (2010) | 11 economic time series of the NN5 Forecasting Competition | A new Jordan RNN based on hybrid complex NN was applied and compared to MLP and multi-recurrent NN. |
| | [51] (2011) | Flash oods without rainfalls | Comparison between Jordan RNN and MLP to forecast Flash oods without rainfalls. |
| | [5] (2012) | Wind speed | Comparison between two architectures of Jordan RNN and SARIMA method. |
| | [52] (2012) | Solar Radiation | A variation of Jordan RNN was proposed called Wavelet RNN and applied to forecast Solar Radiation. |
| | [19] (2013) | Electricity | A variation of Jordan RNN called: Recurrent Cartesian Genetic Programming evolved ANN was applied to forecast electrical load. |

### 3.1. Function approximation and forecasting values based on REESNN

A Jordan RNN (JRNN) is a neural network that uses the received previous network output as a new input to be later processed [54]. Hence, a JRNN is a simple recurrent network so that a speci c group of neurons (later called context neurons) receives feedback signals from the previous time step. Thereby, input layer is composed of two parts: true input neurons and context neurons which are a duplicate of the outputs provided from the previous time step. Hence, the structure of the JRNN model can be pictorial as illustrated in Figure 2.

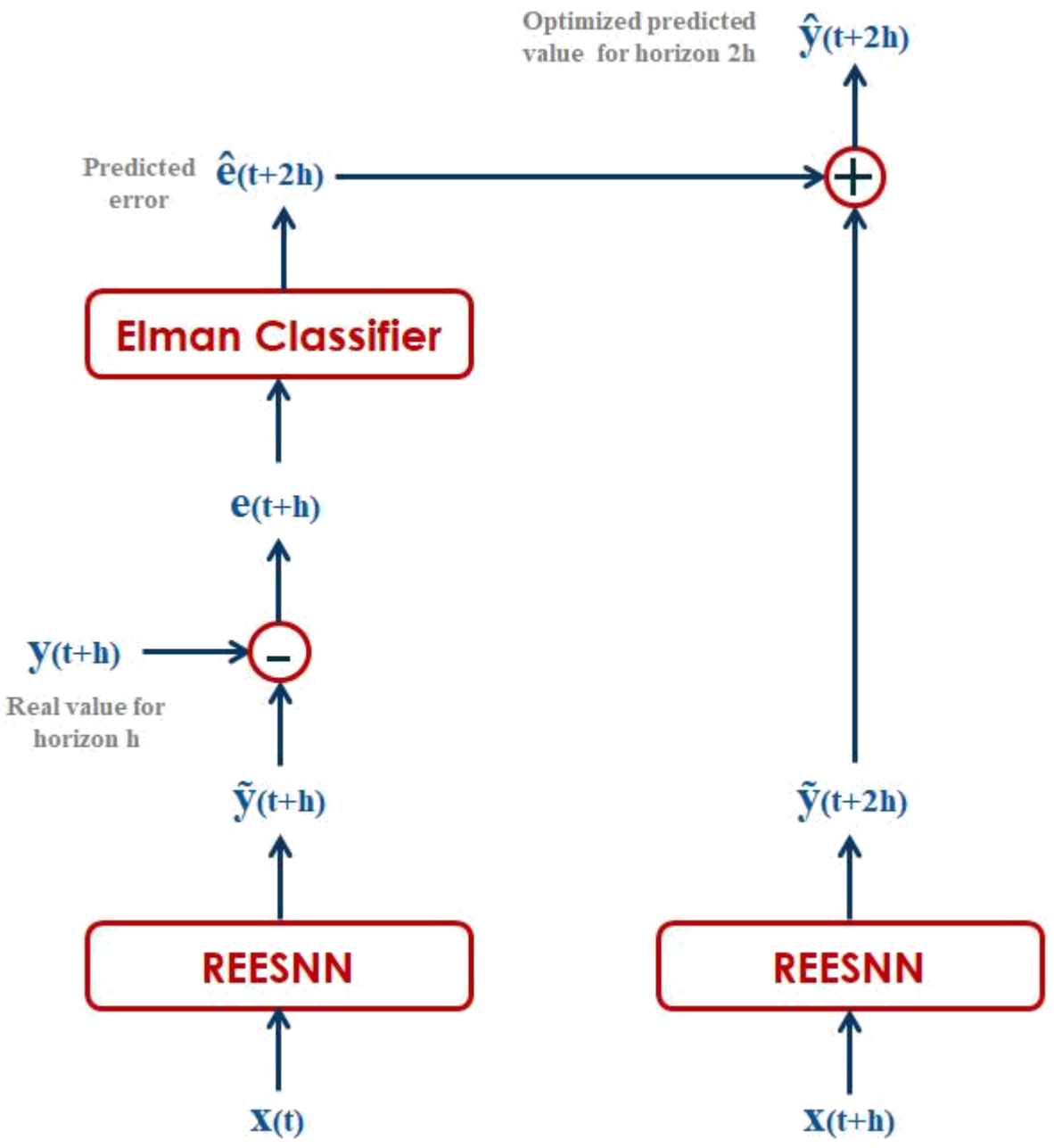

Figure 3: The owchart of a general TOREESNN architecture

Basically, each unit in a particular layer is connected with all neurons in the next layer and then summed and multiplied by the appropriate importance degree called $w_{ij}$ which characterizes the connection and the degree of importance between the $i^{th}$ and the $j^{th}$ neuron. Hence, the nonlinear mapping function F masters the amplitude of the resulting output Y which is given by Equation 1.

$$Y_i = F\left(\sum_{j=1}^{i} w_{ij} h_j\right) \qquad (1)$$

where F is called the activation function. In general, the transfer function accurately re ects the nonlinearity degree of most preprocessed data by NN. We note that in theory, the activation function can be any di erentiable function but in practice only four transfer functions are de ned and can be computed according to Equations 2- 5.

- The linear function:

$$F_{(x)} = x \tag{2}$$

- The sigmoid bipolar function (hyperbolic)

$$F_{(x)} = \frac{e^{x} \quad e^{x}}{e^{x} + e^{x}} \tag{3}$$

-The cos-sin function:

$$F_{(x)} = \sin(x) \text{or} F_{(x)} = \cos(x) \tag{4}$$

- The sigmoid function:

$$F_{(x)} = \frac{1}{1 + e^{cx}} \tag{5}$$

where c is the adaptive gain parameter of sigmoid function.

Meanwhile, in JRNN structure, the activation of the $j^{th}$ hidden neuron $h_j$ is computed by summing the activation of context units in addition to those of true inputs. It holds that:

$$h_j(t) = F\left(\sum_{i=1}^{k} w_{ij}x_i(t) + \sum_{m=1}^{n} v_{mj}y_i(t \quad 1)\right) \tag{6}$$

where $w_{ij}$ is the connection strength between the $j^{th}$ hidden unit $h_j$ and the $i^{th}$ real input $x_i$ , and $v_m j$ is the connection's weight between the $j^{th}$ hidden unit $h_j$ and the $i_t h$ output past value at time t 1. Based on the modeled architecture in 2, the JRNN structure with activation function (5) in both hidden layer and output layer can be expressed by Equation 7.

$$y = \sum_{h=1}^{H} u_h\left[\sum_{i=1}^{k} w_{ij}x_i(t) + \sum_{m=1}^{n} v_{mj}y_i(t \quad 1)\right] \tag{7}$$

where y is the predicted output and uh are the connections weights on output layer.

Subsequently, all connection weights in the network are are merely calculated and updated by applying the back-propagation learning algorithm, as expressed in Equation 8,

$$w_{ij} = \frac{\partial E}{\partial w_{ij}} \tag{8}$$

where is the learning rate and E is the estimated error between the actual observation $y_i$ and its estimated value $\hat{y}_i$.

$$E = \frac{1}{2} \sum_{i=1}^{n} (y_i \;\; \hat{y}_i) \tag{9}$$

Hence, our goal is to understate the error E in order to adjust the network's connection weights and update them according to (10).

$$w_{ij}(t + 1) = w_{ij}(t) + \; w_{ij} \tag{10}$$

In the same way, all connection strengths in JRNN architecture are so updated. In our study, to approximate the value of a function, we considered recursive architecture inspired from the Exponential Smoothed (ES) method [35] which is a statistical forecasting technique. Its main principle assumes that each observation at time t relies on the previous observation and the variation between that forecast and the actual value of the series at that point. The principle of the ES can be formulated by the equation:

$$y(t + 1) = y(t) + \; (x(t) \;\; y(t)) \tag{11}$$

where y(t) is the previous forecasted value , x is the actual one and $\in [0; 1]$ is the smoothing factor.

Our experiments aim to convert the ES method into an Extended ES de-signed as a Jordan Recurrent neural architecture. This task is accomplished within two steps: First, the ES method is modi ed and extended so that each observation does not only depend on solely one past value as ES assumes, but on many past values. Hence, we consider Extended ES (EES) as a generalization case of ES. Mathematically, the equation of EES can be formulated as:

$$y(t + 1) = \sum_{k=0}^{n} (y(t \;\; k) + \; (x(t \;\; k) \;\; y(t \;\; k))) \tag{12}$$

The second step consists on conceiving the EES as a recurrent architecture since it adopts properly the same principle; i.e the prediction of the future forecasted value depends on the previous estimated output as the JRNN supposes. Moreover, the smoothing factor is supposed been replaced by the weight con-nection. Hence, the simplest method for this purpose is to project the EES dynamic equation into a mapping function that can be depicted as:

$$y(t+1) = F\,[y(t); :::; y(t \quad k); (x(t) \quad y(t)); :::; (x(t \quad k) \quad y(t \quad k))] \tag{13}$$

Graphically, the Recursive Extended Exponential Smoothed Neural Net-works (REESNN) is a JRNN where the input layer is a set of two types of inputs: the previous forecasted values y at time t 1; t 2; ::; t k and the estimated error e outlined as the variance between the actual real value x and the forecasted one y at time t 1; t 2; ::; t k as illustrated in Figure 4.

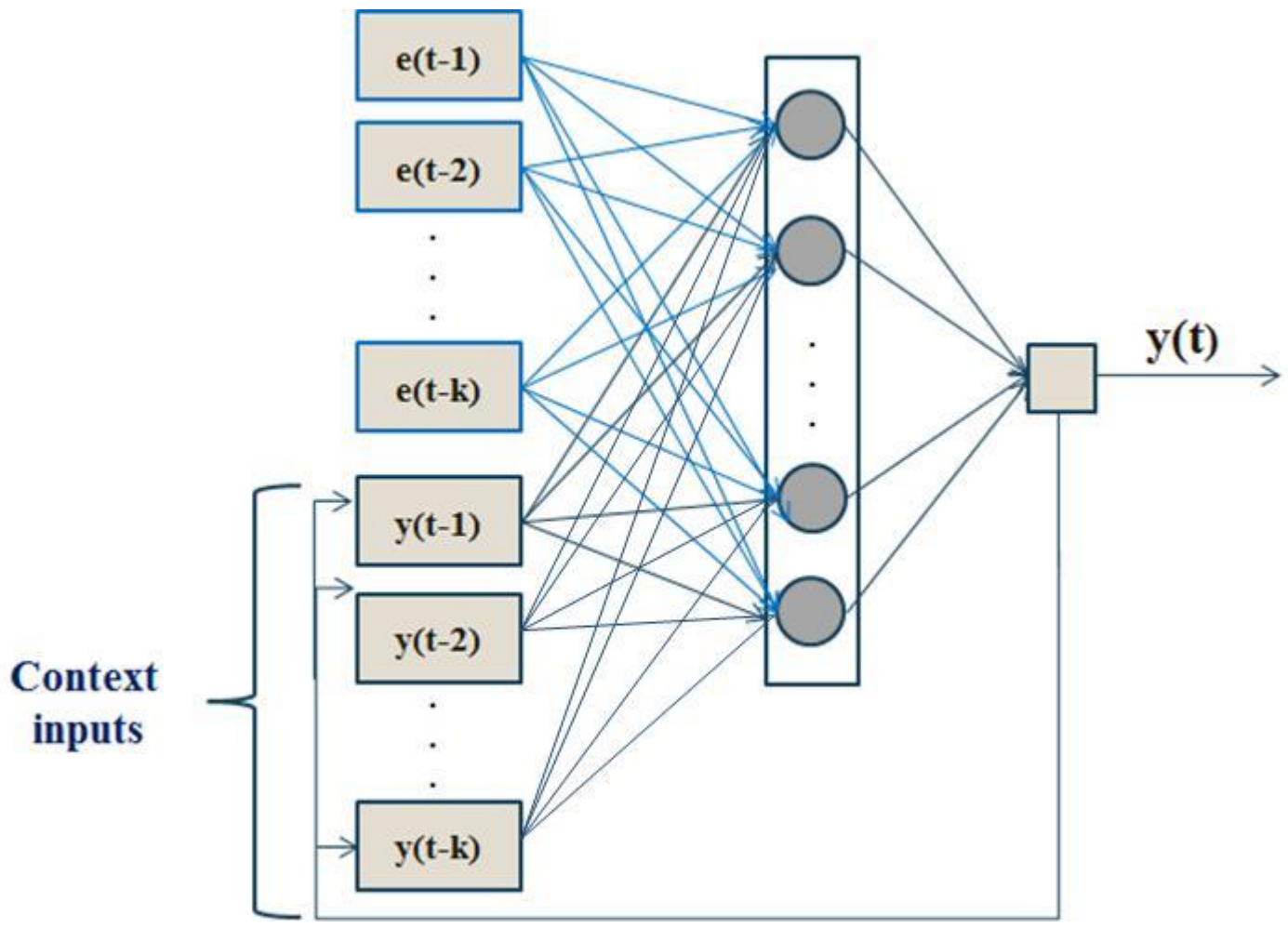

Figure 4: Architecture of Recursive Extended Exponential Smoothed Neural Networks (REESNN).

### 3.2. Forecasting error estimation

A powerful part of this approach is the prediction of error estimation in order to improve the forecasting precision. Forecasting error estimation is de ned as the prediction of the difference between the actual value x and the predicted one y.

However, the uncertainty associated with forecasting is one of the most powerful factors influencing the resulting uncertainty. This point inspires us to examine in advance the task of forecasting by predicting not only the future values but also the uncertainty associated with each value.

In our study, the Elman RNN Classifier (ERNNC) [40] model is used to approximate the error estimation associated to each forecasted value as it has one of the best learning rates when compared to other existing functions in classification forecasting problems.

The suggested model is a three layered NN in addition to a hidden context layer and a classification function as shown in Figure 5.

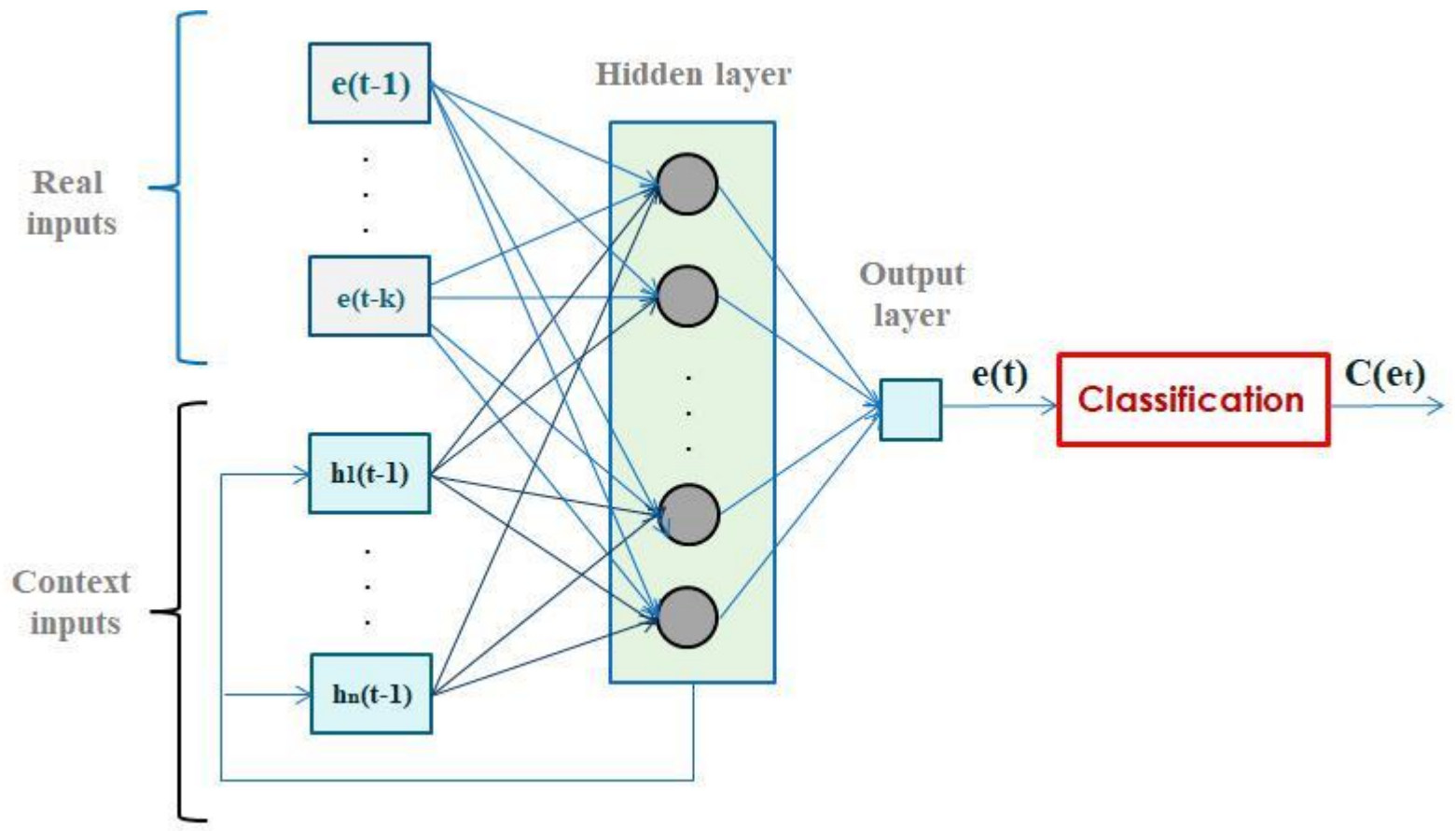

Figure 5: Elman Recurrent Neural Network Classifier architecture.

Firstly, to train the Elman RNN as a Classifier method, it is necessary to preprocess the data and transform it on the adequate form. Hence, the transformation of the training Dataset $S_c$ for the Elman RNN Classifier is introduced as follows:

$$S_c = \{(E_t; C(e))\} \in [0; 1] \times \{-1; 0; 1\} \qquad (14)$$

where $E_t$ is an input vector and $C(e)$ is a classi cation function of the esti-mated error $e$ de ned by Equations (15) and (16).

$$E_t = [e(t\ \ ); e(t\ 2\ ); :::; e(t\ d\ )] \tag{15}$$

$$C(e) = \begin{cases} 1 & e < \\ 0 & e \\ +1 & e > \end{cases} \tag{16}$$

where is the time-delay estimated errors into three scales: positive values, negative values and values near to zero. We note that is chosen, graphically, from the estimated error histogram as Figure 6 depicts.

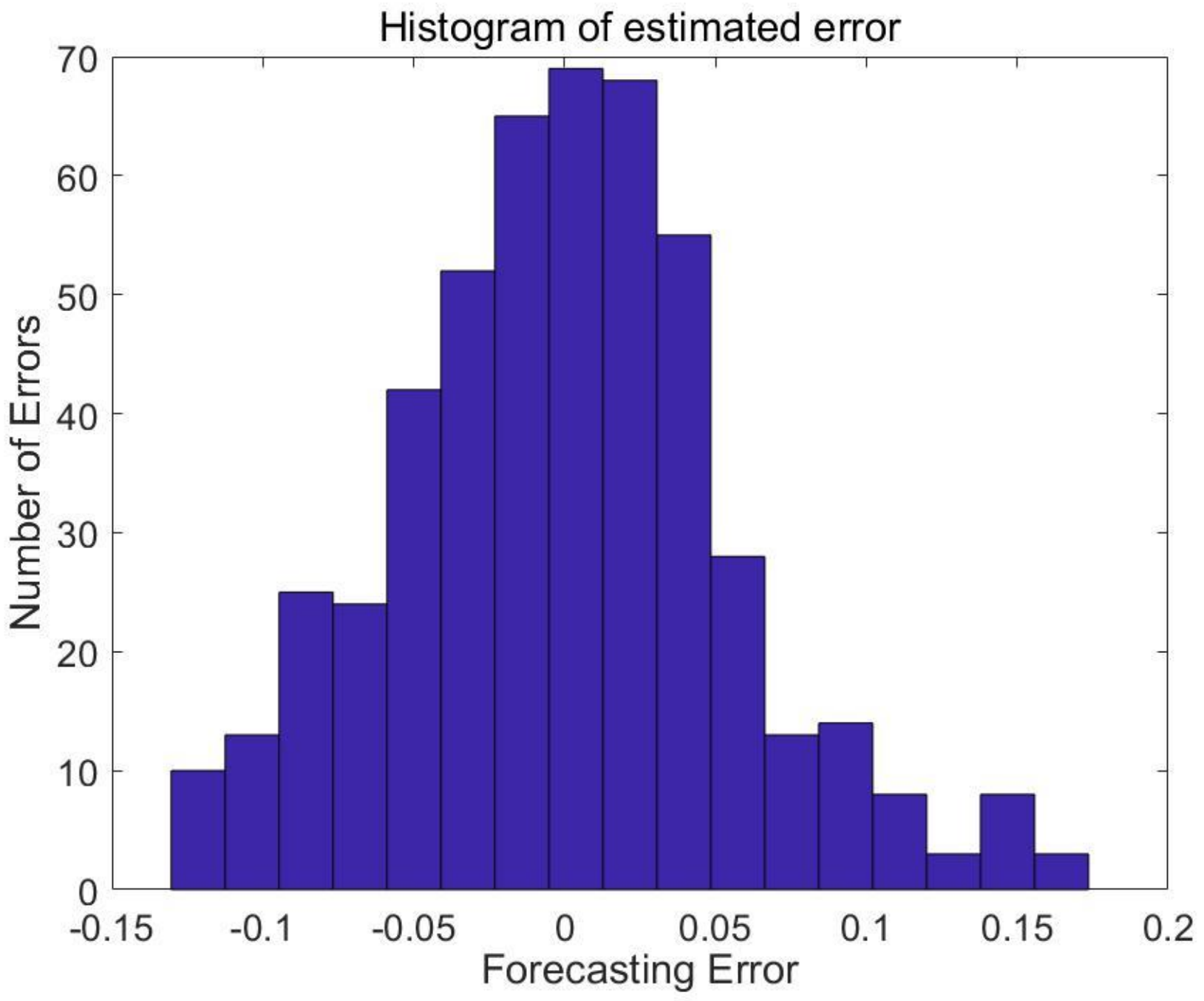

Figure 6: Histogram of estimated error.

After the histogram of the estimated error was established, the Elman RNN Classi er is trained using $S_c$ to predict the error class associated to each sample in the vector $X_t$ .Then, to estimate the forecasting error, IF-THEN rules are adopted.

$$\begin{cases} \text{IF } C(e) = 1 \text{ THEN } f(e) = +! \\ \text{IF } C(e) = 0 \text{ THEN } f(e) = 0 \\ \text{IF } C(e) = 1 \text{ THEN } f(e) = ! \end{cases} \tag{17}$$

where ! is an experimental value near to and f(e) is the estimated forecasting error.

### 3.3. Taylor-based forecasting results optimization

After predicting the estimated error, a step of optimization results was done. The optimized forecasted future value $y_{t+1}$ is calculated according to an approximation function expressed between the first forecasted value $y_t$ resulting from the first part of our architecture denoted as REESNN and the estimated forecasted error f(e) resulting from the second part denoted as Elman Classifier (see Figure 3). Though, a Taylor series approximation represents a number as a polynomial that has a very similar value to the number in a neighborhood around a specified value. Hence, the key idea in our method is to optimize the forecasted results serving by the Taylor expansion [53].

The main principle of the Taylor expansion [53] aims to approximate a function that is many times differentiable in the neighborhood of a point x. Its equation is expressed as:

$$f(x) = f(x_0) + \frac{x - x_0}{1!} f'(x_0) + \frac{(x - x_0)^2}{2!} f^2(x_0) + \dots + \frac{(x - x_0)^n}{n!} f^n(x_0) + (x - x_0)^n \epsilon(x - x_0) \tag{18}$$

where $\epsilon(x - x_0)$ is a function that gets closer and closer to 0 as $(x - x_0)$ tends to 0.

From the above definition, this formula looks pretty similar to the optimization formulation; i.e the key idea behind optimization is to approximate a number by representing it by its nearest accurate neighborhood. Hence, a new problem formularization is proposed which translates the Taylor expansion formula into a combination of optimization task.

In fact, to resolve $y_{t+1}$ and $y_t$ by Equation 18 a step of integration of the error estimation function f(e) is needed before time t+ 1. This integration expression shows that the proposed optimization model can perform the dynamic backward optimization to better exploit the historical knowledge in the studied time series.

Inspired from Taylor principle, and provided the is small enough, the previous equation denotes that this combinational optimization problem can be approximately formulated as:

$$y_{t+1} = y_t + \frac{(t + 1) \;\; t}{1!} + 0(t) \quad (19)$$

where $y_{t+1}$ denotes the optimized forecasted value and $y_t$ represents the predicted value before optimization. We note that in our study we will limit to the rst stage of Taylor principle and we hypothesize that the error approximation f(e) is equal to $y_t^0$ as given in the following equation:

$$y_t^0 = \frac{y_{t+1} \;\; y_t}{(t + 1) \;\; t} \quad (20)$$

## 4. Experimental results

To validate the e ciency of our proposed method, we have based on three data sets. The rst training data is composed of 500 samples which represent 35% from the whole data. The second training data is composed of 500 another patterns to train the estimated error which represents the second part of our architecture. Finally 500 new samples are used as testing data in order to improve the performance of our proposed scheme with new cases.

### 4.1. Mackey-Glass time series

The performance of the proposed methodology is demonstrated via the well-known benchmark time series called the Mackey-Glass time series (MG) [55].

Mathematically, the MG formula is established by a simple dynamic equation that can be noted as:

$$x(n \quad 1) = \frac{1}{1+b}\left[x(n \quad 1) + \frac{ax(n \quad)}{1+x^{10}(n \quad)}\right] \tag{21}$$

where a and b are constants, and  represents the time delay.

In this paper, the used parameters a and b are set to 0:2 and 0:1 respectively, = 17 and time step size $t_0$ = 0:1s. The aim of this paper is to forecast the actual value x(t) based on some speci c historical past values. We note that a step of normalization was done to optimize the training phase.

In order to prove the e ciency of our proposed architecture, di erent forecasting RNN models were compared and testing results are outlined in table 3.

According to table 3, it is revealed that the proposed (TOREESNN) performs better than other RNN methods. Moreover, it can be noticed that optimizing forecasting task by predicting estimated errors in uences greatly the testing results.

Figure 7 illustrates the superposition between our results and the corresponding desired outputs.

Table 3: COMPARISON OF DIFFERENTS FORECASTING RNN MODEL TO OUR PROPOSED METHOD

| Methods | MSE test |
|---|---|
| Elman [39] | 3.00e-03 |
| Elman & Elman Classi er | 2.30e-03 |
| ESN-LARS[56] | 3.20e-02 |
| ESN-FSR[56] | 7.72e-03 |
| ESN-LASSO[56] | 4.31e-03 |
| REESNN | 2.07e-04 |
| TOREESNN | 1.91e-04 |

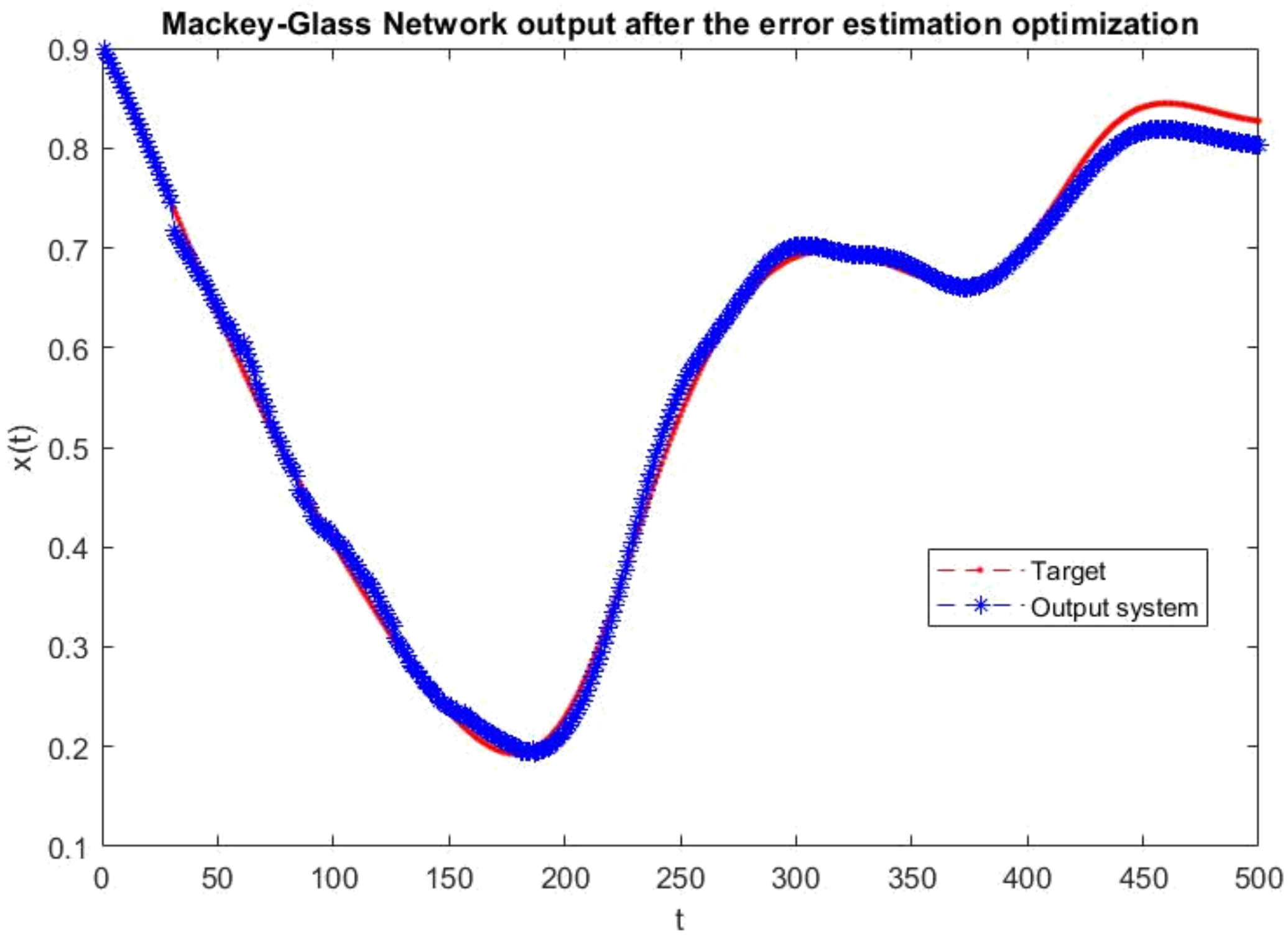

Figure 7: Mackey-Glass Network output after the error estimation optimization

## 4.2. The Nonlinear Auto-Regressive Moving Average time series

The Nonlinear Auto-Regressive Moving Average later called NARMA [57] is a renowned benchmark which is differentiated by its chaotic behavior. Further, the non-correlation between its values makes the task of learning more difficult to achieve. Moreover, the dynamic equation of NARMA is mathematically-dependent on many parameters which makes it hard to model. In light of these facts, NARMA is considered among the most complex studied benchmarks; this point motivates us to further improve the efficiency of our method by testing NARMA patterns.

Mathematically, the NARMA formula is expressed as:

$$y(t+1) = c_1 y(t) + c_2 y(t) \sum_{i=1}^{k} y(t-i) + c_3 x(t-k-1)x(t) + c_4 \qquad (22)$$

where x(t) represents the input, k characterizes the order of the time series and the parameters $c_i$ are set to 0:3, 0:05, 1:5 and 0:1 respectively.

The calculated errors between the output values and the desired one are visualized in gure 8. Seen that the error signal is displayed around 0:05 it clearly re ects that the network response is miming the required one. By the same token as for the foregoing benchmark, a MSE based comparison study with other existent forecasting methods is applied for NARMA series prediction with order k = 10.

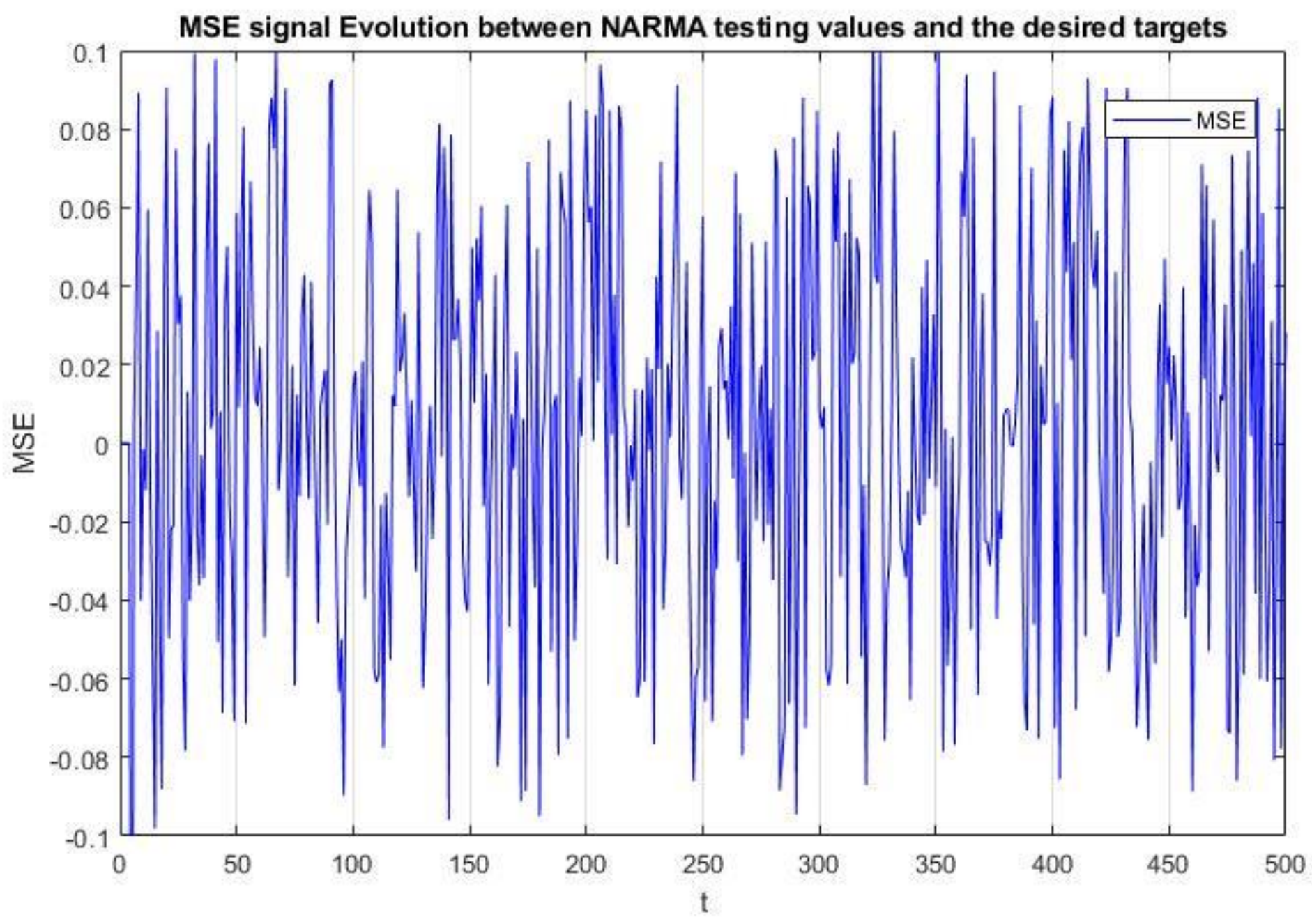

Figure 8: Evolution of the MSE signal between NARMA testing values and the desired targets.

As illustrated by the preceding test, the output system and the target values are plotted in the same gure. The proportionality between the two curves

Table 4: MSE BASED COMPARISON BETWEEN OUR PROPOSED APPROACHE AND OTHER LITERATURE METHODS

| Methods | MSE test |
|---|---|
| ESN-FSR[56] | 7.29e-03 |
| ESN-LARS[56] | 3.79e-03 |
| ESN-RR[56] | 3.03e-03 |
| ESN-LASSO[56] | 2.89e-03 |
| REESNN | 1.20e-02 |
| TOREESNN | 2.00e-03 |

further justifies the obtained results.

### 4.3. Lorenz Attractor

Lorenz can be defined as a system of three nonlinear differential equations characterizing a fluid motion between a hot and a cool surface.(23-25) [58][59].

$$\frac{dx}{dt} = \sigma(y - x) \tag{23}$$

$$\frac{dy}{dt} = -y - xz + rx \tag{24}$$

$$\frac{dz}{dt} = xy - bz \tag{25}$$

where $\sigma = 10$, $r = 28$ and $b = 8/3$.

As in the previous time series, our task is to forecast $x(t)$ according to a set of its own historical past values.

In the same way as the MG test, the resulting tests are displayed in the same figure with the target values to be then compared. The superposition of the two curves is depicted in figure 9. It can be noticeable the network outputs are mimicking the output line of the target values.

As revealed by the previous benchmarks, a MSE based comparison with some existent literature RNN methods is applied for Lorenz attractor. Table 5 recapitulates testing results and proves that our proposed method represents a strong competitive forecaster in terms of accuracy.

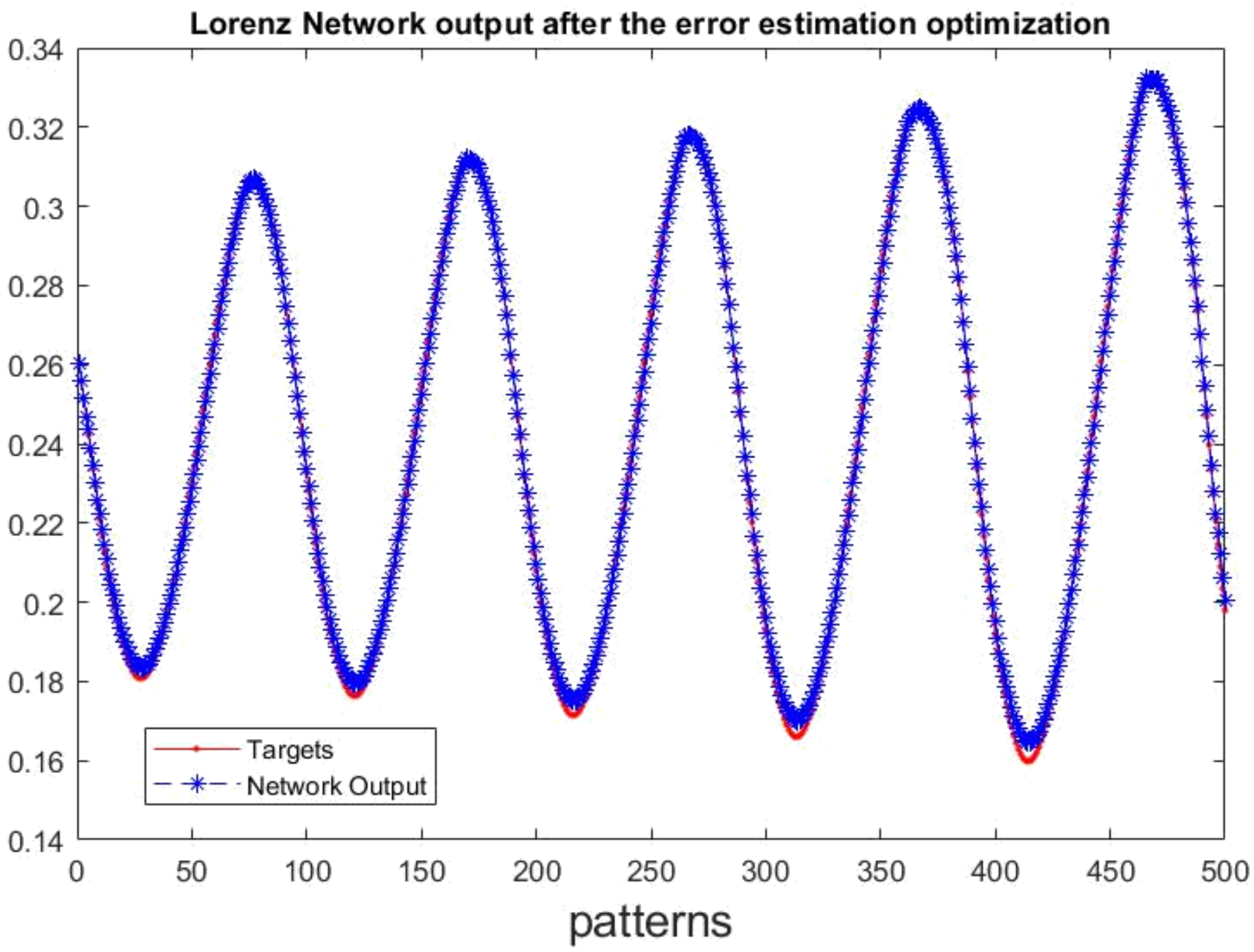

Figure 9: Lorenz Network output after the error estimation optimization

Table 5: MSE BASED COMPARISON BETWEEN OUR PROPOSED APPROACHE AND OTHER LITERATURE METHODS (LORENZ BENCHMARK)

| Methods | MSE test |
|---|---|
| SVM | 2.00e-03 |
| SVR | 2.80e-01 |
| Canonical ESN [60] | 7.23e-04 |
| PSO-ESN [60] | 4.04 e-06 |
| REESNN | 3.93e-06 |
| TOREESNN | 3.90e-06 |

### 4.4. Henon Attractor

The Henon map is a discrete-time dynamic attractor. Litteraturaly, this model is considered among the most common studied dynamical systems. It is typified by its chaotic behavior defined concretely by the following equations in 26 and 27 [58].

$$x(t+1) = y(t) - ax^2(t) + 1 \tag{26}$$

$$y(t+1) = bx(t) \tag{27}$$

where $a = 1:4$ and $b = 0:3$.The dataset was normalized in the interval [0; 1] to get more accurate results during the implementation.

Figure 10 presents a depiction of the MSE signal calculated between desired values and system outputs. According to the figure, the signal of calculated errors is displayed around 0:04 which extremely reflects that the network response is following the desired one.

Table 6 presents the comparison of testing results given by the already existent approaches in the literature for Henon attractor forecasting. According to Table 6, we remark that TOREESNN outperformances the other works in terms of accuracy.

Table 6: MSE BASED COMPARISON BETWEEN OUR PROPOSED APPROACHE AND OTHER LITERATURE METHODS (HENON BENCHMARK)

| Methods | MSE test |
|---|---|
| Canonical ESN [60] | 3.40e-3 |
| SGD-ESN[61] | 3.67e-2 |
| RLDDE[62] | 4.70e-3 |
| REESNN | 9.70e-04 |
| TOREESNN | 8.23e-04 |

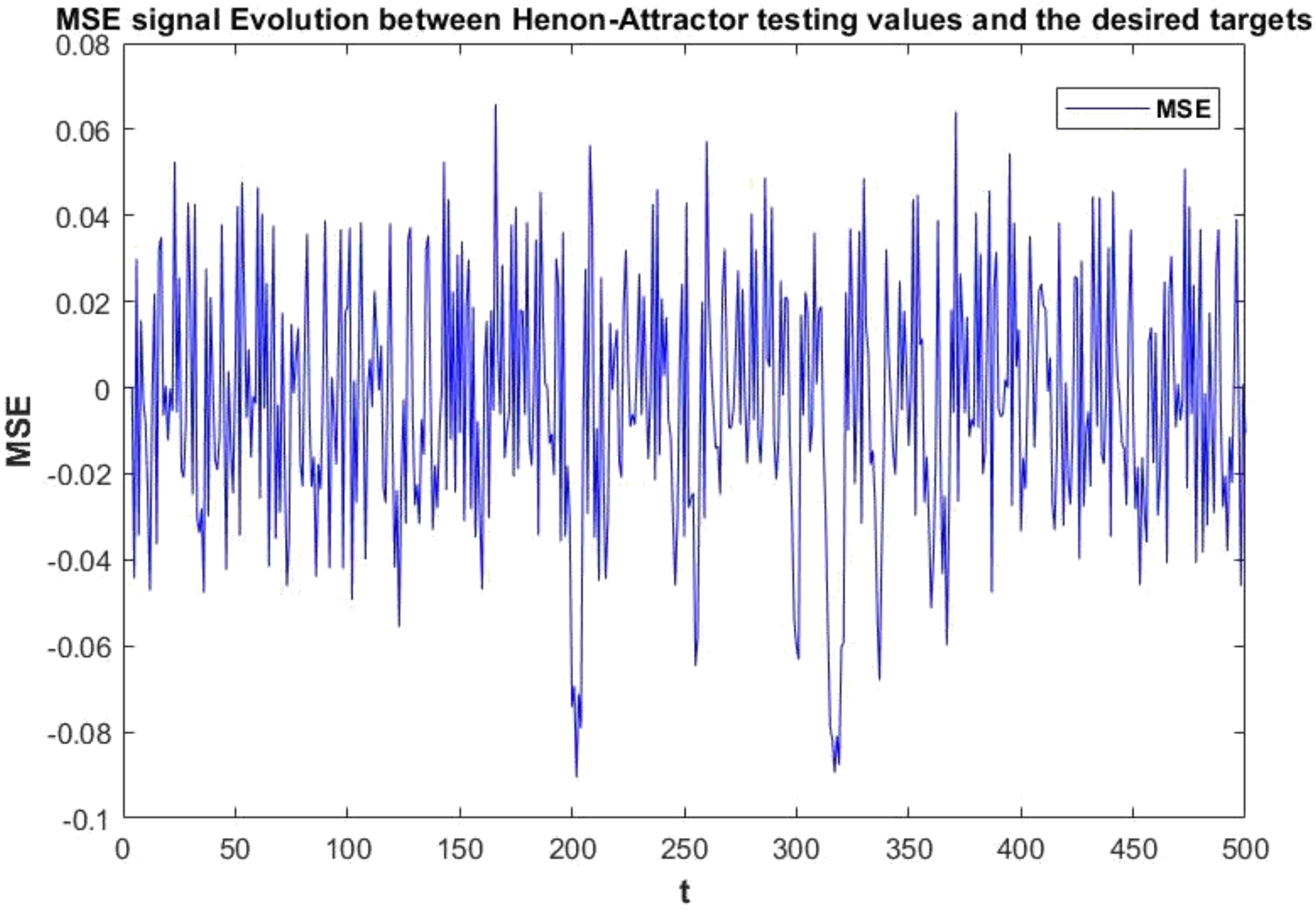

Figure 10: Evolution of the MSE signal between Henon-Attractor testing values and the desired targets.

## 5. Conclusion

In literature, the task of Forecasting was investigated with di erent Arti cial Neural Network methods inter alia Recurrent Neural Networks architectures. In this paper, we applied a novel JRNN architecture (named TOREESNN) inspired from the ES method extended by error approximation to predict and optimize one of the most well-known time series such as Mackey-Glass time series, Henon Attractor, Lorenz and NARMA. We have compared the obtained results to other RNN architectures and experimental results proved the e ciency of our proposed method.

In summary, this paper warrants the assumption above-mentioned at the beginning of this paper which assumes that optimizing the task of forecasting by prediction the error estimation can be a good alternative model to deal hard time series. In fact, the step of optimization o ers the possibility to make full use of historical knowledge in past data as it suits ideally for memorizing not only past forecasted values but also the error made during the training phase.

## Acknowledgements

The research leading to these results has received funding from the Ministry of Higher Education and Scienti c Research of Tunisia under the grant agreement number LR11ES48.